\documentclass{article}

\usepackage{arxiv}

\usepackage{graphicx}

\usepackage[utf8]{inputenc} 
\usepackage[T1]{fontenc} 

\usepackage{graphicx}
\usepackage{subfigure}
\usepackage{amssymb,amsmath}
\usepackage{url}
\usepackage{rotating}
\usepackage{makecell}

\usepackage{balance}
\usepackage[table]{xcolor}
\usepackage{colortbl}
\usepackage{adjustbox} 
\usepackage{tabularx} 
\usepackage{ltablex} 
\usepackage{multirow}
\usepackage{hhline}

\usepackage{fancyhdr,epsfig,psfrag,graphicx,tabularx}
\usepackage[table]{xcolor}
\usepackage{hhline}
\usepackage{array}
\newcolumntype{H}{>{\setbox0=\hbox\bgroup}c<{\egroup}@{}} 
\newcolumntype{C}[1]{>{\centering\arraybackslash}m{#1}} 

\usepackage{booktabs}
\usepackage{url}

\usepackage{algorithm}
\usepackage{algpseudocode,setspace}%

\usepackage{enumitem}
\setlist{leftmargin=12pt}

\newcommand{\citep}{\cite}
\newcommand{\citet}{\cite}

\title{Deep learning models for representing out-of-vocabulary words\thanks{We gratefully acknowledge the support provided by the São Paulo Research Foundation (FAPESP; grants $\#$2017/09387-6, $\#$2018/02146-6), CAPES, and CNPq.}}

\author{Johannes V. Lochter\\ 
Department of Systems and Energy,\\ University of Campinas (UNICAMP),\\ Campinas, S\~{a}o Paulo, Brazil\\ \\
Smart Campus,\\ Engineering College of Sorocaba (Facens),\\ Sorocaba, S\~{a}o Paulo, Brazil\\
\texttt{johannes.lochter@facens.br} \\
\And
Renato M. Silva, Tiago A. Almeida\\
Department of Computer Science, \\Federal University of S\~{a}o Carlos (UFSCar), \\Sorocaba, S\~{a}o Paulo, Brazil\\
\texttt{renatoms@dt.fee.unicamp.br}, \texttt{talmeida@ufscar.br} \\
} 

\begin{document}
\maketitle

\begin{abstract}

Communication has become increasingly dynamic with the popularization of social networks and applications that allow people to express themselves and communicate instantly. In this scenario, distributed representation models have their quality impacted by new words that appear frequently or that are derived from spelling errors. These words that are unknown by the models, known as out-of-vocabulary (OOV) words, need to be properly handled to not degrade the quality of the natural language processing (NLP) applications, which depend on the appropriate vector representation of the texts. To better understand this problem and finding the best techniques to handle OOV words, in this study, we present a comprehensive performance evaluation of deep learning models for representing OOV words. We performed an intrinsic evaluation using a benchmark dataset and an extrinsic evaluation using different NLP tasks: text categorization, named entity recognition, and part-of-speech tagging.  
Although the results indicated that the best technique for handling OOV words is different for each task, Comick, a deep learning method that infers the embedding based on the context and the morphological structure of the OOV word, obtained promising results.

\end{abstract}

\section{Introduction}
\label{sec:introduction}

As the world became more digital, the amount of unstructured data available on the Internet has increased to the point where it has become unbearable to handle it using human resources. Natural language processing (NLP) tasks were developed to address it in an automatic way using computational resources. Some examples of these tasks are audio transcription, translation, assessment on text summaries, grading tests, and opinion mining~\cite{Cho:2014}.

For NLP tasks, a critical point is the computational text representation since there is no consensus on how to represent text properly using computational resources. The most classical text representation is bag-of-words. In this distributive representation, a vocabulary is collected in the training corpus and each sample\footnote{In this study, we use the word sample to denote instance or text document.} is represented by a vector where each element represents the occurrence or absence (1/0) of vocabulary terms in the document \cite{Lochter:2018}.

New text representation techniques have been studied due to known issues of bag-of-words representation: it loses word locality and fails to capture semantic and syntactic features of the words. To address these issues, other techniques were developed, such as the distributed text representation that learns fixed-length vector for each word, known as word embeddings. Using statistics of the context and the abundant occurrences of the words in the training corpus, learned word embeddings can capture the semantic similarity between words~\cite{mikolov:2013-word2vec}. 

As each sample can have many words, a composition function is usually employed to encode all word embeddings into a single fixed-length representation per sample to satisfy the fixed-length input restriction on the most of the predictive models. Some representation techniques also encodes the position of a word in the sample, addressing the word locality issue~\cite{Cho:2014}.

The majority of predictive models for NLP tasks have their performance degraded when unknown words, which were not collected to build the vocabulary in the training phase or were discarded due to low frequency across the corpus, appear in the test. These words are called out-of-vocabulary (OOV) words and can degrade the performance of NLP applications due to the inefficiency of representation models to properly learn a representation for them. 

In order to emphasize how an OOV word can hinder sentence comprehension, consider the following example originally written in ``The Jabberwocky'' by Lewis Caroll: ``He went galumphing back''. The nonce word ``galumphing'' was coined to mean ``moving in a clumsy, ponderous, or noisy manner; inelegant''. Since this word is an OOV, traditional models are not capable to handle it properly, ignoring it. The lack of representation for this word can restrict the  predictive model capabilities to understand this sentence~\cite{adams:2017}.

Handling OOV words in distributed representation models can be achieved with simple strategies. For instance, as OOV words have no word embedding representation learned, they can be ignored when the composition function is applied. This approach leads the predictive model to fit data without the knowledge of the absence of a word because it is unknown to the representation model. For such case, a random vector can be adopted for each OOV word or a single random vector can be adopted for all OOV words~\cite{yang_2018_uwetec}. 

These simple strategies provide little or no information about unknown words to predictive models in downstream tasks. In order to enable a predictive model to use a vector representation for the unknown words, those words need to be replaced by a meaningful in-vocabulary word. For this specific task, most of the techniques available in literature fits in two groups: language models \cite{Sundermeyer:2012} and robust techniques capable of learning meaningful representation for OOV words using their structures or the context in which they appear~\cite{garneau:2018-pieoovdt,bojanowski:2017-fasttext}.

In these two groups, there are several deep learning (DL) models.  Some of them were developed to handle OOV, such as Comick~\cite{garneau:2018-pieoovdt} and HiCE~\cite{hu:2019_hice}, while evidence was found that pure neural architectures can also perform it, such as LSTM~\cite{Mikolov:2010_b} and Transformer~\cite{Vaswani:2017}. Some language models also had success in this task, such as RoBERTa~\cite{Liu:2019}, DistillBERT~\cite{Sanh:2019}, and Electra~\cite{Clark:2020}.

Although several studies have shown that DL can be successfully applied in several NLP tasks, such as sentiment analysis~\cite{Ouyang:2015}, named entity recognition (NER), and part-of-speech (POS) tagging~\cite{hu:2019_hice}, there are few DL models for handling OOV words and no consensus on which approach is the best. To fill that gap, in this paper, we present a performance evaluation of state-of-the-art DL models  considering different datasets and tasks that can be greatly affected by OOV words.

\section{OOV Handling}
\label{sec:oov_handling}

When a word in a sample is unknown to a representation model, this word is an OOV word to the model. If the representation model is unable to properly handle OOV while generating a vector to represent a sentence or document, the OOV is ignored and no information about it is added to the vector.  This lack of information tends to degrade the performance of the predictive models as the number of OOV words per sample increases.

Simple replacement methods are straight forward solution to OOV handling, such as the replacement of every OOV word by the same random vector or a different random vector for each OOV word.

There is also the zero-vector representation replacement which is suitable to inhibit activation through the neural network for that specific OOV word~\cite{yang_2018_uwetec}. More elaborated methods for OOV handling also showed good results, although they were not able to learn new representations. For instance, Khodak \textit{et al.}~\cite{khodak:2018} proposed to average the embeddings of the closest words of a OOV word in the sample. Their proposed method failed to capture complex semantic relationship, but obtained better performance than ignoring the OOV words. 

Some other methods are capable of handling OOV words by learning good enough representation to keep or enhance predictive model performance using OOV morphological structure and context information. In addition, some methods are able to predict the word of the vocabulary that is most similar to the OOV using a language model.

\subsection{Approaches based on the word context or structure}

According to Hu \textit{et al.}~\cite{hu:2019_hice}, the representation methods able to learning new representations to handle OOV usually employ two different sources of information: the morphological structure of the OOV words~\cite{pinter:2017,bojanowski:2017-fasttext}, or the context in which the OOV is inserted~\cite{garneau:2018-pieoovdt,hu:2019_hice}.

\subsubsubsection{\textbf{FastText~\cite{bojanowski:2017-fasttext}}} A popular distributed representation model that associates the morphological structure (subword) to the vector representation of the words. This feature enables FastText to handle OOV or rare words using its morphological learning capabilities. In this representation model, each word representation is composed as a bag of character n-grams in addition to the word itself. Considering the word `model' with n = 3, fastText n-gram representation is <mo, mod, ode, del, el>. Symbols < and > are added as boundary to distinguish n-grams from a word itself. When a small word is already in the vocabulary, \textit{e.g.} `mod', it is represented as <mod> to preserve its meaning. Such approach is capable of capture meaning for suffixes and prefixes and find a representation for OOV or rare words using vector composition for each word.

\subsubsubsection{\textbf{Mimick~\cite{pinter:2017}}} Like FastText, also uses the word structure to represent an OOV. It is a deep learning model that uses every character of the OOV word to produce a vector representation. A neural network is fed with the characters of the word, and the target is a known representation for that word in a trained representation model. The objective function of the neural network minimizes the distance between the characters of the word and its known representation in the representation model. After Mimick is trained, a vector representation for an OOV word can be estimated using its characters.

The main drawback of methods that use only morphological information is their incapacity to handle OOV words that have different meanings in different contexts. When morphological information is the only source to generate a representation, an OOV word will always have the same representation regardless
of the context in which it appears. To address it, a group of methods identified by Hu \textit{et al.}~\cite{hu:2019_hice} learns representation for an OOV word using context information, instead only morphological information.

\begin{figure}[h]
\centering
\label{fig:comick}
\includegraphics[width=12cm]{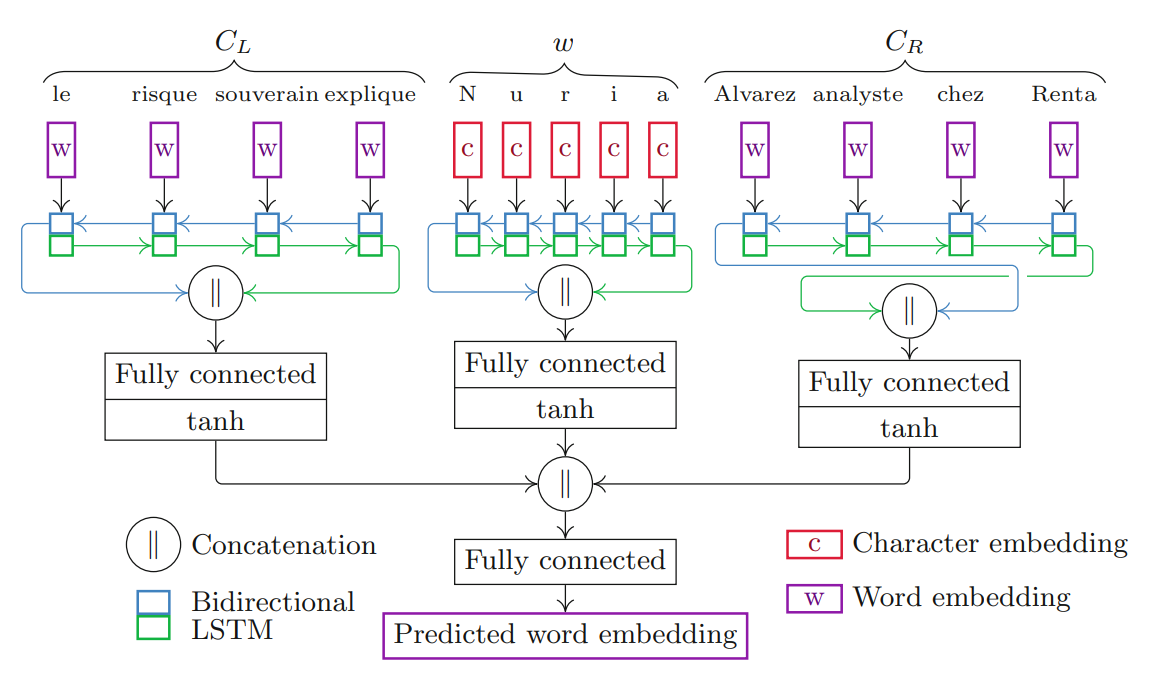}
\caption{Comick architecture~\cite{garneau:2018-pieoovdt}.}
\end{figure}

\subsubsubsection{\textbf{Comick~\cite{garneau:2018-pieoovdt}}} It uses Mimick architecture as reference, but also consider context information around OOV to find a good representation. The architecture of Comick is shown in Figure~\ref{fig:comick}. The OOV handling on Comick is split in three parts: left context, OOV itself, and right context. Each part performs a vector composition using a Bi-LSTM architecture, but both contexts feed Bi-LSTM with word embeddings, while OOV feeds with character embedding. All of three parts are fully-connected to dense layers using hyperbolic tangent activation function.

Their experimental setup was conducted in part-of-speech tagging task in three different languages (English, Spanish, and French) on two different ways: using random embeddings to every OOV and using Mimick obtained. They also compared Comick results to Mimick ones, noticing out improvements across all languages assessed.

\subsubsubsection{\textbf{HiCE~\cite{hu:2019_hice}}} Similar to Comick, this approach has capabilities of learn a representation for an OOV word using both morphological information and context using a deep architecture that finds a good enough representation for a word using the concept of few-shot learning. Using only few examples to feed a hierarchical attention network, HiCE found good enough representations for rare words, according to the high degree of confidence and the high Spearman correlation reported by Hu \textit{et al.} \cite{hu:2019_hice}. The authors also proposed an adaptation model to extend the learned representations to different domains.

\begin{figure}[h]
\centering
\label{fig:hice}
\includegraphics[width=8cm]{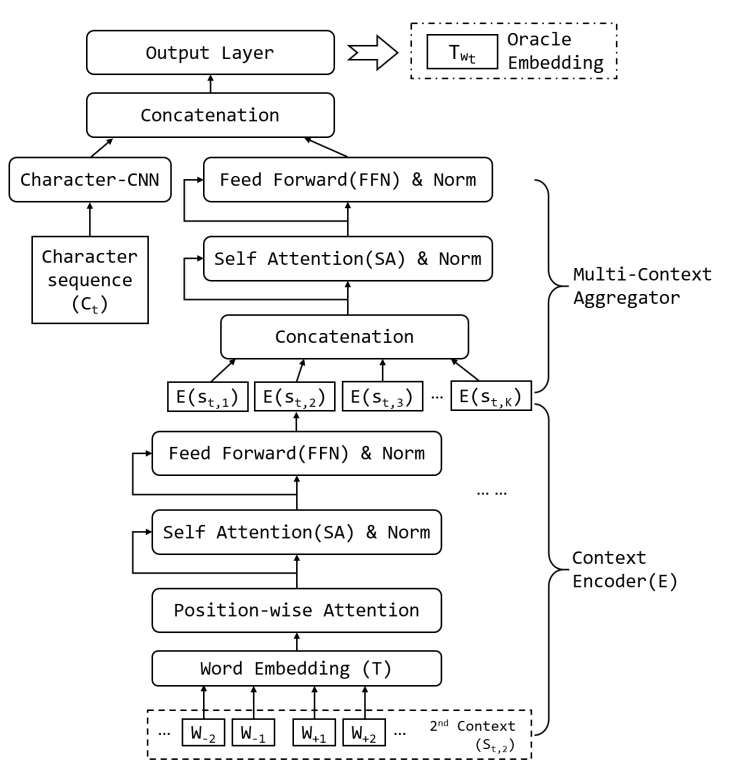}
\caption{HiCE architecture~\cite{hu:2019_hice}.}
\end{figure}

The architecture proposed by authors (Figure~\ref{fig:hice}) is composed of three blocks: context encoder, aggregator, and character-cnn. The first block, described  as context encoder is fed with words around OOV in the original sentence. Their original word embeddings are put into an attention mechanism. For HiCE, contexts may be larger than two, as usual in other architectures, because they are left and right across few sentences. The second block is aggregator, which is responsible to apply another attention mechanism to combine all context encoded from first block. Finally, the result is obtained from a concatenation across aggregator and character-cnn, which is the third block. This block is a vanilla CNN to combine character embeddings from the OOV itself.

\subsection{Deep learning based language models}

Some language models are capable of predict the next word in a sentence while preserving the semantics of the original context due to their ability to learn semantic features from corpora~\cite{Chomsky:1957}. While most traditional language models are statistical models, recent approaches uses deep learning (DL) architectures to perform neural network language model task with higher accuracy~\cite{Sundermeyer:2012}. In the following, several DL language models are described.

\vspace{0.1cm}
\noindent
\textbf{LSTM.} Recurrent neural networks are appropriate for sequence modeling due to their memory capability, which makes it possible to take into account all predecessor words, instead of a fixed context length as in feed-forward networks~\cite{Mikolov:2010_b}. This architecture is hard to train because it suffers with the vanishing gradient problem while propagating gradient through the network. LSTM was proposed as a solution to the vanishing gradient problem using gates to define which features should be remembered across the network with a scaling factor fixed to one. The results obtained by Sundermeyer \textit{et al.}~\cite{Sundermeyer:2012} indicated that LSTM is better for sequence modeling in recognition systems than back-off models, which were considered
state-of-the-art at the time.

\vspace{0.1cm}
\noindent
\textbf{Transformer.} Firstly introduced in Vaswani \textit{et al.}~\cite{Vaswani:2017} as an alternative to complex recurrent or convolutional neural networks in an encoder-decoder configuration, the transformer architecture is a simple network architecture based solely on attention mechanisms, dispensing with recurrence and convolutions entirely. As opposed to LSTM, which reads the text input sequentially (left-to-right or right-to-left), the transformer architecture reads the entire sequence of words at once, which allows the model to learn the context of a word based on all of its surroundings. It is also efficient in terms of computational cost as it is more parallelizable, and it achieved state-of-the-art performance in many NLP tasks.

\vspace{0.2cm}

After the transformer architecture became avaliable, many works were proposed using it as cornerstone (\textit{e.g.}, GPT-2 and BERT family). They use the principle of solely rely on attention mechanism for language modeling.

\vspace{0.1cm}
\noindent
\textbf{GPT-2.} Defined by authors as ``large transformer-based language model''~\cite{Radford:2019}, it was trained on a dataset of 8 million web pages to predict the next word, given all of the previous words within the text. The authors claim that GPT-2 is able to perform tasks as zero-shot, which means a task can be performed without any data of the specific domain.

\vspace{0.2cm}

Previously efforts on using transformer to language modeling relied on looking a text sequence from left-to-right or combined directions left-to-right and right-to-left in training phase due to the task of predicting next word. BERT (Bidirectional Encoder Representations from Transformers)~\cite{Devlin:2018} can have a deeper sense of language context and flow than single-direction language models because is bidirectionally trained. To make it possible, the authors of BERT redefined task on training phase as (1) masked model language (MLM), where instead to predict the next word, the model should predict any word masked (replaced by [MASK]) in the sample, and (2) predicting next sequence, where the model receives pairs of sentences as input and learns to predict if the second sentence is the subsequent one to the first sentence in the original document.

It is still an open challenge to train large models, such as BERT, and make them feasible for inference since their size is prohibitive in terms of computational cost. Considering that, many efforts have been reported in the literature to improve these models lowering the minimum requirements to run them, such as RoBERTa~\cite{Liu:2019}, DistillBERT~\cite{Sanh:2019}, and Electra~\cite{Clark:2020}.

\vspace{0.1cm}
\noindent
\textbf{RoBERTa.} Although BERT has shown state-of-the-art improvements in many NLP tasks in the last years, so far, it is expensive to process large corpus to train. RoBERTa~\cite{Liu:2019} is a new approach to pretrain on BERT in a optimized way. Instead of a new architecture, RoBERTa is described as a new model which removes next sentence pretraining objective from original BERT and uses larger mini batches and learning rates to improve performance across many NLP tasks. The effectiveness of this new model is also related to a new dataset using public news article several times bigger than the one employed on BERT training.

\vspace{0.1cm}
\noindent
\textbf{DistillBERT.} As described by Sanh \textit{et al.}~\cite{Sanh:2019}, DistillBERT is a smaller general-purpose language representation model which can be fine-tuned on specific domain tasks in NLP. On their work, the authors claim that knowledge distillation on pretraining can reduce BERT size by 40\%, keeping 97\% of language understanding capabilities, producing a model 60\% faster.

\vspace{0.1cm}
\noindent
\textbf{Electra.} While masked language models require a large amount of computational resources to train a model due to the expensive process of reconstruct original tokens which were masked, Electra~\cite{Clark:2020} is an alternative more efficient that uses a replacement process with likely tokens instead the masking process from original BERT work to train the model. The results claimed by authors indicate lower costs on training with huge improvements to small models, leading to even better results than originally observed with BERT model. The authors also claim that for large models, such as RoBERTa, it uses less than 25\% of computational resources, and can outperform them using the same amount of resources.

\vspace{0.2cm}
In this study, we present a performance evaluation of Comick, HiCE, LSTM, Transformer, GPT-2, RoBERTa, DistillBERT, and Electra in the task of handling OOV words. As presented above, these DL models have different strategies for handling OOVs. Therefore, a comparative analysis between them, in addition to revealing the best strategy, may present important insights for future research on the problem of handling OOV words. In the following, we present the experiments performed to evaluate these models. We performed an intrinsic evaluation using a benchmark dataset and an extrinsic evaluation using different NLP tasks. 

\section{Intrinsic evaluation}
\label{sec:intrinsic_evaluation}

To analyze the ability of DL models to find good representations for OOVs, we first performed an intrinsic evaluation using the benchmark Chimera dataset \cite{lazaridou:2013} that simulates OOV words in real-world applications. In this dataset, two, four, or six sentences are used to determine the meaning of a OOV, called chimera. This OOV is a nonword used to simulate a new word whose meaning is a combination of the meanings of two existing words. For each chimera, the dataset provides a set of six probing words and the human annotated similarities between the probing words and the chimeras \cite{lazaridou:2013}. Figure \ref{fig:example_chimera} presents an example of a chimera from the Chimera dataset.

\begin{figure}[!htb]
\centering
\footnotesize

\def\arraystretch{0.8}
\begin{tabular}{p{0.90\textwidth}}
\toprule
\textbf{Chimera:} pirbin (a combination of \textbf{alligator} and \textbf{rattlesnake}) \\ \midrule
\textbf{Sentence 1:} \sffamily The outside section will also be used by PIRBINS, the rare L'hoest and Diana monkeys, cheetah and leopards. \\
\textbf{Sentence 2:} \sffamily But the kangaroo rat can hear the faint rustles of the PIRBIN's scales moving over the sand, and escape. \\
\textbf{Sentence 3:} \sffamily Blackmer and Culp are by now halfway across the swamp and have attracted the attention of several PIRBINS.\\
\textbf{Sentence 4:} \sffamily Faced with jewels, I sort of did a story and put a jungle PIRBINS into it. \\ \midrule
\textbf{Probe words}: crocodile; iguana; gorilla; banner; buzzard; shovel \\ \midrule
\textbf{Human annotated similarities:} crocodile (2.29); iguana (3.43); gorilla~(2.14); banner (1.57); buzzard (2.71); shovel(1.43)  \\
\bottomrule
\end{tabular}

\caption{Example of a Chimera defined by four sentences.}
\label{fig:example_chimera}
\end{figure}

 Table \ref{tab:dataset_Chimera} presents the main statistics about the Chimera dataset, where $|C|$ is the number of chimeras, and $|V|$ is the number of unique terms (vocabulary). Moreover,  $\mathcal{M}^{t}$ and $\mathcal{I}^{t}$ are the median and the interquartile range of the number of terms per sentence.

 \begin{table}[!htb]
 	\centering
 	\footnotesize
 	\caption{Basic statistics about the Chimera dataset.}\label{tab:dataset_Chimera}
 	\setlength{\tabcolsep}{9.5pt} 
 	\begin{tabular}{lllll}
 		\toprule
 		  & $C$   & $|V|$ & $\mathcal{M}^{t}$ & $\mathcal{I}^{t}$ \\ \midrule
 		2 sentences & 330  & 3,663 & 17 & 5 \\
 		4 sentences        & 330 & 5,845 & 17 & 5 \\
 		6 sentences  & 330    & 7,711 & 18 & 5 \\ \midrule
 	\end{tabular}
 \end{table}

To generate the vector representation of the sentences of the Chimera dataset, we used FastText word embeddings~\cite{bojanowski:2017-fasttext} trained on Twitter7 (T7) \cite{Yang:2011}, a corpus of tweets posted from June 2009 to December 2009. We removed the retweets and empty messages of this corpus and selected only the English-language messages. At the end, we used the remaining 364,025,273 tweets to train the embeddings. 

In this study, we infer an embedding for a given Chimera based on each sentence (two, four, or six sentences) that forms its meaning. If the model is unable to infer a vector for the chimera in a given sentence, the embedding of that chimera in that sentence is a vector of zeros. The final embedding for the chimera is the average of the vectors obtained for each sentence. 

For each Chimera, we calculated the cosine similarity between the embedding for the chimera and the embedding of each of the probe words. Then, to evaluate the performance of a model for a given chimera, we calculated the Spearman correlation between the cosine similarities obtained by the model and the human annotations, as performed in other related studies (\textit{e.g.,}  Hu \textit{et al.~}\cite{hu:2019_hice} and Khodak \textit{et al.~}\cite{khodak:2018}). In experiments where the model was unable to infer the embedding for a given chimera in any sentences, resulting in a zero vector for the embedding, we assigned a value of zero for the correlation (worst possible value). The overall performance of the model is the average Spearman correlation across all chimeras.

\subsection{Baseline methods}

We compare the results of the DL models with the following baselines:
\begin{itemize}[topsep=2.5pt]
    \item \textbf{Oracle}: it is probably the best possible embedding for the Chimera, since it is the average vector of the embeddings of the two gold-standard words that compose the Chimera. Therefore, we considered the Spearman correlation between the results obtained by the Oracle and the human annotations as the upper bound performance.
    
    \item \textbf{Sum}: the vector of the Chimera in a sentence is the sum of the words embeddings of this sentence. Then, the final vector for the Chimera, is the average vector of the sentences that form the Chimera.
    
    \item \textbf{Average}: the vector of the Chimera in a sentence is the average of the embeddings of the words of this sentence. Then, the final vector for the Chimera, is the average vector of the sentences that form the Chimera.
\end{itemize}

\subsection{Deep learning models}
\label{sec:deeplearningModels_settings}

A chimera is a word that does not exist in the real-world. 
Therefore, in the intrinsic evaluation, it was not possible to perform experiments with techniques that analyze the morphological structure of the word, such as Comick. 
We present below the experimental settings for the evaluated DL models :
\begin{itemize}[topsep=2.5pt]
    \item \textbf{DistilBERT}: it was used the original model available in the \texttt{transformers}\footnote{Transformers. Available at https://huggingface.co/transformers, accessed on \today.} library for Python 3, identified by the keyword \texttt{`distilbert-base-cased'}.
    \item \textbf{HiCE (context)}: this model was trained using 10\% of the T7 dataset using the default parameters of the official implementation\footnote{HiCE. Available at https://github.com/acbull/HiCE, accessed on \today.}. However, the morphological information flag was set to \texttt{false}, since the OOVs in this experiment (chimeras) are not real-world words and, therefore, we can use only its context to infer its embedding.
    \item \textbf{GPT2}: it was used the original model available in the \texttt{transformers} library for Python 3, identified by the keyword \texttt{`gpt2-large'}.
    \item \textbf{Electra}: this language model was trained using 10\% of the T7 dataset using the default parameters, except for the number of epochs, which was set to 10.  The implementation for Electra model is available in the \texttt{transformers} library for Python~3.
    \item \textbf{LSTM}: this model was trained using 10\% of the T7 dataset using default parameters, except for the number of epochs, which was set to 10. The implementation is available in the PyTorch\footnote{PyTorch Github. Available at https://bit.ly/2B7LS3U, accessed on \today.} repository.
    \item \textbf{Transformer}: this model was trained using 10\% of T7 dataset using default parameters, except for the number of epochs, which was set to 10.  The implementation for Transformer architecture is available in the \texttt{transformers} library for Python 3.
    \item \textbf{RoBERTa}: it was used the original model available in the \texttt{transformers} library for Python 3, identified by the keyword \texttt{`roberta-base'}.
\end{itemize}

For all DL models, we used the default parameters because the computational cost to make an appropriate parameter selection is very high. 

As the embeddings returned by the DL models are not in the same vector space of the word embeddings trained on T7, the models return a list of five candidate words to replace the OOV. The vector of the first candidate in the T7 vocabulary is used to represent the OOV.

\subsection{Results}
\label{sec:results_chimera}

Table \ref{tab:results_chimera} shows the average Spearman correlation obtained on the Chimera dataset. The average ranking of each method is also presented. For each experiment, the method that obtained the best average Spearman correlation received rank 1 and the worst one obtained rank
9 (nine techniques were evaluated). Therefore, the smaller the average ranking, the better the performance. The
results are presented as a grayscale heat map, where the better the value, the darker the cell color. Bold values indicate the best result. Moreover, the methods are sorted by the average ranking.

To complement the analysis of the results, Table \ref{tab:results_chimeras_stat} presents the percentage of sentences where the evaluated DL model was able to find a vector for the chimera. The baseline methods (average and sum) always return a vector because they do not make any kind of prediction and, therefore, it was not necessary to include them in the table.

\begin{table}
\centering
\scriptsize
\setlength{\tabcolsep}{4.0pt} 
\caption{Results obtained in the intrinsic evaluation.}%
\label{tab:results_statistics_chimeras}	
\subtable[][\label{tab:results_chimera}Average Spearman correlation.]{
\begin{tabular}{C{0.5cm}lcccc} 

\toprule

&  & \multicolumn{3}{c}{Avg. correlation} & \multirow{3}{*}{\makecell{Avg. \\ranking}} \\ \cmidrule(lr){3-5}

&  & \makecell{2\\sent.} & \makecell{4\\sent.} & \makecell{6\\sent.} & \\ \cmidrule(lr){3-5}  \cmidrule(lr){6-6}

 & Oracle                                 & 0.40 & 0.40 & 0.43 & - - \\ \cmidrule(lr){3-5} \cmidrule(lr){6-6}

\multirow{2}{*}{\rotatebox{90}{\makecell{Base-\\lines}}} 
& Average                  & \cellcolor{gray!45}0.26          & \cellcolor{gray!50}0.29          & \cellcolor{gray!45}0.30          & 2.00          \\
 & Sum                      & \cellcolor{gray!45}0.26          & \cellcolor{gray!45}0.28          & \cellcolor{gray!45}0.30          & 2.33                   \\
\cmidrule(lr){3-5} \cmidrule(lr){6-6} 

\multirow{7}{*}{\rotatebox{90}{\makecell{Deep\\learning}}} 
 & DistilBERT               & \cellcolor{gray!60}\textbf{0.27}          & \cellcolor{gray!60}\textbf{0.31}          & \cellcolor{gray!60}\textbf{0.37}          & 1.00          \\

 & HiCE (context)           & \cellcolor{gray!43}0.16          & \cellcolor{gray!43}0.26          & \cellcolor{gray!43}0.29          & 4.00          \\
 & GPT2                     & \cellcolor{gray!37}0.15          & \cellcolor{gray!37}0.22          & \cellcolor{gray!37}0.20          & 5.00          \\
 & Electra                  & \cellcolor{gray!31}0.06          & \cellcolor{gray!31}0.17          & \cellcolor{gray!31}0.16          & 6.00          \\
 & LSTM                     & \cellcolor{gray!31}0.06          & \cellcolor{gray!26}0.11          & \cellcolor{gray!20}0.10          & 6.67          \\
 & Transformer              & \cellcolor{gray!20}0.05          & \cellcolor{gray!20}0.04          & \cellcolor{gray!20}0.10          & 7.67          \\
 & RoBERTa                  & \cellcolor{gray!15}0.03          & \cellcolor{gray!15}0.03          & \cellcolor{gray!15}0.02          & 9.00          \\

 \midrule

	\end{tabular}}
	\qquad
\subtable[][\label{tab:results_chimeras_stat}Statistics of the OOVs.]{
\begin{tabular}{llccc}
\toprule

& & \makecell{2\\sent.} & \makecell{4\\sent.} & \makecell{6\\sent.} \\ \midrule 

\multirow{7}{*}{\rotatebox{90}{\makecell{\% of OOVs \\treated by \\the models}}} & \makecell{HiCE (context) }                        & 100.0          & 100.0          & 100.0          \\ 
 & Electra                                & 100.0          & 99.0          & 99.0          \\
& GPT2                                   & 98.0          & 97.0          & 97.0          \\

& LSTM                                   & 97.0          & 97.0          & 97.0          \\
& Transformer                            & 97.0          & 97.0          & 97.0          \\
& DistilBERT                             & 95.0          & 95.0          & 95.0          \\
&  RoBERTa                                & 9.0          & 10.0          & 10.0          \\
\midrule

	\end{tabular}	
}
\end{table}

In general, DistilBERT was the best technique for inferring embeddings for the chimeras. The average Spearman correlation of this tecnique was the closest
to the upper bound performance (the one obtained by the Oracle) on the experiments with two, four, and six sentences. 

The baseline approaches (average and sum) obtained, respectively, the second and third best performance. On the other hand, RoBERTa obtained the lowest Spearman correlation in all experiments, which also resulted in the worst average ranking. These results can be better understood when analyzed together with the statistics shown in Table \ref{tab:results_chimeras_stat}. We can see that RoBERTa was able to infer embeddings for the Chimeras, in at most only 10\% of the sentences.  

HiCE was able to infer embeddings for the chimeras in 100\% of the sentences, but obtained only the fourth best performance, in general. On average, its score was 26\% lower than the one obtained by the best model (DistilBERT).

LSTM, Transformer, and GPT2 use only the context window to the left of the OOV to infer its embedding. Some chimeras are located in the first position of the sentence, leaving no context window to be used by these models. Therefore, in these sentences, they are not able to infer the vector for the chimera. Despite this, the percentage of chimeras handled by these models was still higher than DistilBERT and RoBERTa.

\section{Extrinsic evaluation}
\label{sec:extrinsic_evaluation}

We conduct a comprehensive extrinsic evaluation of the DL models in three established tasks that can be greatly affected by OOV words: 

\begin{itemize}[topsep=2.5pt]
 	\item \textbf{text categorization}: we address the task of polarity sentiment classification in short and noise messages.
	
	\item \textbf{named entity recognition (NER)}: this task seeks to locate and classify entities in a sentence.
	
 	\item \textbf{part-of-speech (POS) tagging}: this task seeks to identify the grammatical group of a given word. 
 \end{itemize}

In all experiments, we used the same FastText word embeddings applied in the intrinsic evaluation to generate the vector representation of the documents.

\subsubsection{Baseline methods}

We compare the results of the DL models with the following baselines:
\begin{itemize}[topsep=2.5pt]

    \item \textbf{Sum}: the OOV is represented by the sum of the embeddings of the words in the document. 
    
    \item \textbf{Average}: the OOV is represented by the average of the embeddings of the words in the document.
    
    \item \textbf{Zero}: the OOV is represented by a vector of zeros.
    
    \item \textbf{Random}: all OOVs are represented by the same random vector generated at the beginning of the experiment. 
    
    \item \textbf{FastText}: the OOV is represented by the vector obtained by the FastText~\cite{bojanowski:2017-fasttext}. 
\end{itemize}

\subsubsection{Deep learning models}

In the extrinsic evaluation, we performed experiments with the same DL methods used in the intrinsic evaluation using the same experimental settings described in Section \ref{sec:deeplearningModels_settings}. Additionally, we also evaluated models that use the morphological information of the OOV word to infer an embedding. We present below the experimental settings for these DL models:

\begin{itemize}[topsep=2.5pt]

    \item \textbf{HiCE}: this model was trained using 10\% of the T7 dataset using the default parameters of the official implementation\footnote{HiCE. Available at https://github.com/acbull/HiCE, accessed on \today.}. However, the morphological information flag was set to \texttt{true}, since, in this experiment, the word structure of the OOV words can be used to predict reliable embeddings.
    \item \textbf{Comick}: this model was trained using 10\% of the T7 dataset using the default parameters on the private implementation obtained from Garneau \textit{et al.}~\cite{garneau:2018-pieoovdt}.
    
\end{itemize}

For all DL models, we used the default parameters because the computational cost to make an appropriate parameter selection is very high. 
Both HiCE and Comick infer the representation for the OOV in the same way as the other DL models we evaluated (see Section \ref{sec:deeplearningModels_settings}). 

\subsection{Text categorization}
\label{sec:metodologia_twitter}

In order to give credibility to the results and make the experiments reproducible, all tests were performed with the following real and public datasets: Archeage, Hobbit, and IPhone6~\citet{Lochter:2016}; 
OMD (Obama-McCain \textit{debate})~\cite{shamma:2009:tducaus}, 
HCR (\textit{health care reform})~\cite{speriosu:2011:tpclpllfg}, 
Sanders \cite{speriosu:2011:tpclpllfg}, 
SS-Tweet (\textit{sentiment strength} Twitter \textit{dataset})~\cite{thelwall:2012:ssdsw}, 
STS-Test (Stanford Twitter \textit{sentiment test set})~\cite{agarwal:2011:satd}, and 
UMICH \cite{Lochter:2016}. 

All sentences were converted to lowercase and they were processed using Ekphrasis~\cite{baziotis:2017_dsst}, a tool to normalize text from social media.

Table~\ref{tab:datasets} presents the main statistical for the datasets, where $|D|_{\text{Pos}}$ and $|D|_{\text{Neg.}}$ are the number of messages with positive and negative polarity, respectively. Moreover, $|V|$ is the number of unique words in the datasets (vocabulary size), while $\mathcal{M}^{t}$ and $\mathcal{I}^{t}$ are the median and the interquartile range of the number of words per message, respectively.

 \begin{table}[!htb]
 	\centering
 	\footnotesize
 	\caption{Basic statistics about the text categorization datasets.}\label{tab:datasets}
 	\setlength{\tabcolsep}{6.5pt} 
 	\begin{tabular}{lHlllll}
 		\toprule
 		Dataset                                                             & $D$   & $|D|_{\text{Pos.}}$ & $|D|_{\text{Neg.}}$ & $|V|$ & $\mathcal{M}^{t}$ & $\mathcal{I}^{t}$ \\ \midrule
 		Archeage & 1,520 & 591 & 929 & 2,952 & 17 & 12 \\
 		HCR & 1,406 & 534 & 872 & 4,076 & 25 & 9 \\
 		Hobbit & 488 & 327 & 161 & 1,302 & 15 & 12 \\
 		IPhone6 & 522 & 365 & 157 & 1,548 & 15 & 14 \\
 		OMD & 1,886 & 702 & 1,184 & 3,835 & 17 & 9 \\
 		Sanders & 1,078 & 515 & 563 & 3,027 & 20 & 11 \\
 		SS-Tweet & 2,288 & 1,252 & 1,036 & 6,782 & 17 & 10 \\
 		STS-Test & 359 & 182 & 177 & 1,580 & 14 & 12 \\
 		UMICH & 1,116 & 617 & 499 & 2,129 & 10 & 9 \\
 		\midrule
 	\end{tabular}
 \end{table}

\subsubsection{Evaluation}

The experiments were carried out with a Bidirectional LSTM. We built the LSTM using Keras\footnote{Keras. Available at \url{https://keras.io/}. Accessed
	on \today.} on top of TensorFlow\footnote{TensorFlow. Available at \url{https://www.tensorflow.org/}. Accessed
	on \today.}. All the documents were padded or truncated to 200 words. The OOVs that were not handled and paddings words were represented by a vector of zeros. We did not perform any parameter optimization for the neural network  because the objective of this study is not to obtain the best possible result, but to analyze the ability of DL models to handle OOVs. The experiments were performed using stratified holdout validation with 70\% of the documents in the training set and 30\% in the test set. To compare the results, we employed the macro F-measure.

\subsubsection{Results}
\label{sec:results_twitter}

Table \ref{tab:results_twitter} presents the macro F-measure. The average ranking of each method is also presented. For each dataset, the method that obtained the best macro F-measure received rank 1 and the worst one obtained rank
14 (fourteen methods were evaluated). The methods are
sorted by the average ranking. Bold values indicate the
best scores. Moreover, the scores are presented as a grayscale heat map, where
the better the score, the darker the cell color.

To complement the analysis of the results, Table \ref{tab:results_twitter_stat}  presents the percentage of documents that have some OOV. Moreover, among the documents that have some OOV, Table \ref{tab:results_twitter_stat} shows the percentage of them that had at least one OOV handled. The baseline methods (zero, random, sum, and average) always return a vector and, therefore, it was not necessary to include them in the table.

\begin{table}[!htb]
\centering
\scriptsize
\setlength{\tabcolsep}{6pt} 
\renewcommand*{\arraystretch}{0.95}
\caption{Experiments on text categorization.}%
    \label{tab:results_statistics_twitter}

\subtable[][\label{tab:results_twitter}Macro F-measure  obtained  on text categorization.]{
	\begin{tabular}{llcccccccccc}
		\toprule

& & \multicolumn{9}{c}{Macro F-measure} & \multirow{2}{*}{\makecell{Avg. \\ranking}} \\ \cmidrule(lr){3-11} 
& & Archeage & HCR & Hobbit & IPhone6 & OMD & SS-Tweet & STS-Test & Sanders & UMICH & \\  \cmidrule(lr){3-11}  \cmidrule(lr){12-12}

\multirow{5}{*}{\rotatebox{90}{Baselines}} & Zero                                         & \cellcolor{gray!53}0.84                      & \cellcolor{gray!35}0.64                      & \cellcolor{gray!32}0.88                      & \cellcolor{gray!32}0.71                      & \cellcolor{gray!53}0.79                      & \cellcolor{gray!56}0.79                      & \cellcolor{gray!49}0.90                      & \cellcolor{gray!60}\textbf{0.86}             & \cellcolor{gray!49}0.95                      & 4.78          \\
 & Average                                      & \cellcolor{gray!53}0.84                      & \cellcolor{gray!21}0.62                      & \cellcolor{gray!56}0.91                      & \cellcolor{gray!42}0.72                      & \cellcolor{gray!53}0.79                      & \cellcolor{gray!42}0.77                      & \cellcolor{gray!32}0.89                      & \cellcolor{gray!46}0.85                      & \cellcolor{gray!15}0.94                      & 6.33          \\
 & Sum                                          & \cellcolor{gray!53}0.84                      & \cellcolor{gray!42}0.65                      & \cellcolor{gray!28}0.87                      & \cellcolor{gray!18}0.67                      & \cellcolor{gray!60}\textbf{0.80}             & \cellcolor{gray!15}0.64                      & \cellcolor{gray!32}0.89                      & \cellcolor{gray!60}\textbf{0.86}             & \cellcolor{gray!60}\textbf{0.96}             & 6.44          \\
 & Random                                       & \cellcolor{gray!60}\textbf{0.85}             & \cellcolor{gray!53}0.67                      & \cellcolor{gray!18}0.83                      & \cellcolor{gray!42}0.72                      & \cellcolor{gray!15}0.77                      & \cellcolor{gray!25}0.76                      & \cellcolor{gray!32}0.89                      & \cellcolor{gray!18}0.82                      & \cellcolor{gray!49}0.95                      & 8.00          \\
 & FastText                                     & \cellcolor{gray!21}0.81                      & \cellcolor{gray!60}\textbf{0.68}             & \cellcolor{gray!28}0.87                      & \cellcolor{gray!32}0.71                      & \cellcolor{gray!15}0.77                      & \cellcolor{gray!42}0.77                      & \cellcolor{gray!21}0.88                      & \cellcolor{gray!39}0.84                      & \cellcolor{gray!15}0.94                      & 8.89          \\
  \cmidrule(lr){3-11}  \cmidrule(lr){12-12}

\multirow{9}{*}{\rotatebox{90}{Deep learning}} &  Comick                                       & \cellcolor{gray!53}0.84                      & \cellcolor{gray!60}\textbf{0.68}             & \cellcolor{gray!15}0.80                      & \cellcolor{gray!53}0.74                      & \cellcolor{gray!60}\textbf{0.80}             & \cellcolor{gray!56}0.79                      & \cellcolor{gray!53}0.92                      & \cellcolor{gray!39}0.84                      & \cellcolor{gray!60}\textbf{0.96}             & 3.89          \\
 & Transformer                                  & \cellcolor{gray!60}\textbf{0.85}             & \cellcolor{gray!25}0.63                      & \cellcolor{gray!39}0.89                      & \cellcolor{gray!60}\textbf{0.76}             & \cellcolor{gray!35}0.78                      & \cellcolor{gray!42}0.77                      & \cellcolor{gray!60}\textbf{0.93}             & \cellcolor{gray!39}0.84                      & \cellcolor{gray!60}\textbf{0.96}             & 4.78          \\
 & GPT2                                         & \cellcolor{gray!53}0.84                      & \cellcolor{gray!49}0.66                      & \cellcolor{gray!53}0.90                      & \cellcolor{gray!42}0.72                      & \cellcolor{gray!53}0.79                      & \cellcolor{gray!49}0.78                      & \cellcolor{gray!49}0.90                      & \cellcolor{gray!15}0.77                      & \cellcolor{gray!49}0.95                      & 5.00          \\
 & RoBERTa                                      & \cellcolor{gray!21}0.81                      & \cellcolor{gray!42}0.65                      & \cellcolor{gray!53}0.90                      & \cellcolor{gray!49}0.73                      & \cellcolor{gray!53}0.79                      & \cellcolor{gray!42}0.77                      & \cellcolor{gray!49}0.90                      & \cellcolor{gray!39}0.84                      & \cellcolor{gray!49}0.95                      & 5.44          \\
 & Electra                                      & \cellcolor{gray!35}0.83                      & \cellcolor{gray!35}0.64                      & \cellcolor{gray!21}0.84                      & \cellcolor{gray!32}0.71                      & \cellcolor{gray!53}0.79                      & \cellcolor{gray!42}0.77                      & \cellcolor{gray!49}0.90                      & \cellcolor{gray!60}\textbf{0.86}             & \cellcolor{gray!49}0.95                      & 6.11          \\
 & HiCE                                         & \cellcolor{gray!25}0.82                      & \cellcolor{gray!18}0.61                      & \cellcolor{gray!53}0.90                      & \cellcolor{gray!49}0.73                      & \cellcolor{gray!35}0.78                      & \cellcolor{gray!49}0.78                      & \cellcolor{gray!60}\textbf{0.93}             & \cellcolor{gray!60}\textbf{0.86}             & \cellcolor{gray!15}0.94                      & 6.22          \\
 & HiCE (context)                               & \cellcolor{gray!35}0.83                      & \cellcolor{gray!49}0.66                      & \cellcolor{gray!53}0.90                      & \cellcolor{gray!32}0.71                      & \cellcolor{gray!15}0.77                      & \cellcolor{gray!60}\textbf{0.80}             & \cellcolor{gray!21}0.88                      & \cellcolor{gray!39}0.84                      & \cellcolor{gray!49}0.95                      & 6.67          \\
 & LSTM                                         & \cellcolor{gray!15}0.80                      & \cellcolor{gray!15}0.60                      & \cellcolor{gray!60}\textbf{0.92}             & \cellcolor{gray!56}0.75                      & \cellcolor{gray!35}0.78                      & \cellcolor{gray!21}0.75                      & \cellcolor{gray!49}0.90                      & \cellcolor{gray!39}0.84                      & \cellcolor{gray!15}0.94                      & 8.11          \\
 & DistilBERT                                   & \cellcolor{gray!35}0.83                      & \cellcolor{gray!35}0.64                      & \cellcolor{gray!39}0.89                      & \cellcolor{gray!15}0.64                      & \cellcolor{gray!35}0.78                      & \cellcolor{gray!18}0.72                      & \cellcolor{gray!15}0.86                      & \cellcolor{gray!46}0.85                      & \cellcolor{gray!49}0.95                      & 9.00          \\
 \midrule

	\end{tabular}}
\subtable[][\label{tab:results_twitter_stat}Statistics of the OOVs.]{
\begin{tabular}{c|m{1.9cm}ccccccccc} 

\toprule

\multicolumn{1}{c}{} &  & Archeage & HCR & Hobbit & IPhone6 & OMD & SS-Tweet & STS-Test & Sanders & UMICH \\ \midrule 
\multicolumn{1}{c}{} & \% of docs with OOV & 51.0 & 4.0 & 6.0 & 61.0 & 3.0 & 5.0 & 3.0 & 9.0 & 2.0 \\ \midrule 

\multirow{9}{*}{\rotatebox{90}{\makecell{\% of docs \\treated by \\the models}}} & 

 HiCE                                   & 100.0          & 100.0          & 100.0          & 100.0          & 100.0          & 100.0          & 100.0          & 100.0          & 100.0          \\
& HiCE (context)                         & 100.0          & 100.0          & 100.0          & 100.0          & 100.0          & 100.0          & 100.0          & 100.0          & 100.0          \\
& Comick                                 & 100.0          & 100.0          & 100.0          & 100.0          & 100.0          & 100.0          & 100.0          & 100.0          & 100.0          \\

& Electra                                & 99.0          & 100.0          & 100.0          & 100.0          & 100.0          & 95.0          & 90.0          & 100.0          & 95.0          \\

& GPT2                                   & 91.0          & 97.0          & 93.0          & 95.0          & 88.0          & 93.0          & 90.0          & 93.0          & 79.0          \\

& LSTM                                   & 91.0          & 97.0          & 93.0          & 94.0          & 88.0          & 92.0          & 90.0          & 93.0          & 79.0          \\
& Transformer                            & 91.0          & 97.0          & 93.0          & 94.0          & 88.0          & 92.0          & 90.0          & 93.0          & 79.0          \\

& DistilBERT                             & 75.0          & 45.0          & 76.0          & 89.0          & 75.0          & 65.0          & 80.0          & 70.0          & 79.0          \\
& RoBERTa                                & 27.0          & 43.0          & 34.0          & 17.0          & 39.0          & 33.0          & 20.0          & 39.0          & 47.0          \\

 \midrule

	\end{tabular}}

\end{table}

In general, none of the baseline techniques and DL models excelled in all datasets. Comick obtained the best score in three (HCR, OMD, and UMICH) and the best average ranking, and Transformer obtained the best macro F-measure in four (Archeage, Iphone, STS-Test, and UMICH). Two of these, Iphone6 and Archeage, are the ones that have the highest percentage of documents with OOV. They are also the only datasets with more than 10\% of documents with an OOV. Therefore, they are probably the ones with the greatest degree of difficulty. Furthermore, RoBERTa was the technique that treated the lowest percentage of documents with OOVs. Despite this, RoBERTa obtained the fourth best average ranking among the DL methods.

DistilBERT, the best model in the intrinsic evaluation (Section \ref{sec:results_chimera}), obtained the worst average in the text categorization. One of the factors that may have contributed to the worsening of the result is the amount of OOVs that it was unable to handle. In the intrinsic evaluation, it was able to infer the vector for the chimera in 95\% of the sentences, on average (Table \ref{tab:results_chimeras_stat}). However, on text categorization (Table \ref{tab:results_twitter_stat}), on average, DistilBERT was able to infer vectors for OOVs in only 72\% of the documents that had some OOV. 


As in almost all text categorization datasets, most messages do not have any OOV word, we performed experiments inserting artificial OOVs to better evaluate the OOV handling models. For each message, one, two or three words were randomly selected to be considered OOV words. Then, each selected word was modified through one of the following operations:

\begin{itemize}
	\item \textbf{duplication}: a random number of characters in the word have been duplicated.
	\item \textbf{removal}: a random number of characters in the word have been removed;
	\item \textbf{alteration}: a random number of characters in the word have been changed by other random characters.
\end{itemize}

The number of artificial OOV generated took into account the amount of OOVs that the document already had. For example, in the experiment with three artificial OOVs, if a given message already had two OOVs, only one more OOV was created artificially.

For each word transformed into OOV, the type of operation (by duplication, removal or alteration) was chosen randomly. The number of characters was also chosen at random to be one or two.

Tables \ref{tab:results_twitter_1ArtificialOOV}, \ref{tab:results_twitter_2ArtificialOOV}, and \ref{tab:results_twitter_3ArtificialOOV} present the results obtained in the experiments with one, two, and three artificial OOVs, respectively. Additionally, Tables \ref{tab:results_twitter_1ArtificialOOV_stat}, \ref{tab:results_twitter_2ArtificialOOV_stat}, and \ref{tab:results_twitter_3ArtificialOOV_stat} present the percentage of messages were at leat one OOV was handled by the evaluated DL models in the experiments with one, two, and three artificial OOVs, respectively.

\begin{table}[!htb]
\centering
\scriptsize
\setlength{\tabcolsep}{6pt} 
\renewcommand*{\arraystretch}{0.92}
\caption{Experiments with one artificial OOV in each message.}%
    \label{tab:results_stat_twitter_1artificialOOVs}

\subtable[][\label{tab:results_twitter_1ArtificialOOV}Macro F-measure  obtained  on text categorization.]{
	\begin{tabular}{llcccccccccc}
		\toprule

& & \multicolumn{9}{c}{Macro F-measure} & \multirow{2}{*}{\makecell{Avg. \\ranking}} \\ \cmidrule(lr){3-11} 
& & Archeage & HCR & Hobbit & IPhone6 & OMD & SS-Tweet & STS-Test & Sanders & UMICH & \\  \cmidrule(lr){3-11}  \cmidrule(lr){12-12}

\multirow{5}{*}{\rotatebox{90}{Baselines}}  &

 Average                                      & \cellcolor{gray!42}0.84                      & \cellcolor{gray!49}0.68                      & \cellcolor{gray!42}0.84                      & \cellcolor{gray!46}0.71                      & \cellcolor{gray!53}0.78                      & \cellcolor{gray!35}0.77                      & \cellcolor{gray!60}\textbf{0.95}             & \cellcolor{gray!60}\textbf{0.83}             & \cellcolor{gray!42}0.88                      & 3.44          \\
 & FastText                                     & \cellcolor{gray!56}0.85                      & \cellcolor{gray!15}0.61                      & \cellcolor{gray!32}0.81                      & \cellcolor{gray!28}0.69                      & \cellcolor{gray!60}\textbf{0.79}             & \cellcolor{gray!49}0.78                      & \cellcolor{gray!53}0.91                      & \cellcolor{gray!60}\textbf{0.83}             & \cellcolor{gray!60}\textbf{0.90}             & 4.67          \\
 & Zero                                         & \cellcolor{gray!42}0.84                      & \cellcolor{gray!49}0.68                      & \cellcolor{gray!21}0.78                      & \cellcolor{gray!18}0.66                      & \cellcolor{gray!42}0.77                      & \cellcolor{gray!49}0.78                      & \cellcolor{gray!46}0.89                      & \cellcolor{gray!21}0.80                      & \cellcolor{gray!42}0.88                      & 6.22          \\
 & Random                                       & \cellcolor{gray!60}\textbf{0.86}             & \cellcolor{gray!49}0.68                      & \cellcolor{gray!39}0.83                      & \cellcolor{gray!28}0.69                      & \cellcolor{gray!28}0.76                      & \cellcolor{gray!15}0.75                      & \cellcolor{gray!21}0.84                      & \cellcolor{gray!35}0.82                      & \cellcolor{gray!56}0.89                      & 6.56          \\
 & Sum                                          & \cellcolor{gray!28}0.83                      & \cellcolor{gray!35}0.67                      & \cellcolor{gray!28}0.80                      & \cellcolor{gray!21}0.67                      & \cellcolor{gray!28}0.76                      & \cellcolor{gray!25}0.76                      & \cellcolor{gray!28}0.87                      & \cellcolor{gray!35}0.82                      & \cellcolor{gray!15}0.86                      & 8.89          \\
  \cmidrule(lr){3-11}  \cmidrule(lr){12-12}

\multirow{9}{*}{\rotatebox{90}{Deep learning}}  & 

 Transformer                                  & \cellcolor{gray!42}0.84                      & \cellcolor{gray!49}0.68                      & \cellcolor{gray!15}0.77                      & \cellcolor{gray!53}0.72                      & \cellcolor{gray!28}0.76                      & \cellcolor{gray!49}0.78                      & \cellcolor{gray!35}0.88                      & \cellcolor{gray!35}0.82                      & \cellcolor{gray!42}0.88                      & 5.22          \\
 & LSTM                                         & \cellcolor{gray!28}0.83                      & \cellcolor{gray!25}0.65                      & \cellcolor{gray!35}0.82                      & \cellcolor{gray!53}0.72                      & \cellcolor{gray!60}\textbf{0.79}             & \cellcolor{gray!35}0.77                      & \cellcolor{gray!35}0.88                      & \cellcolor{gray!35}0.82                      & \cellcolor{gray!42}0.88                      & 5.67          \\
 & GPT2                                         & \cellcolor{gray!28}0.83                      & \cellcolor{gray!18}0.63                      & \cellcolor{gray!56}0.87                      & \cellcolor{gray!39}0.70                      & \cellcolor{gray!42}0.77                      & \cellcolor{gray!15}0.75                      & \cellcolor{gray!46}0.89                      & \cellcolor{gray!60}\textbf{0.83}             & \cellcolor{gray!42}0.88                      & 6.00          \\
 & Comick                                       & \cellcolor{gray!56}0.85                      & \cellcolor{gray!49}0.68                      & \cellcolor{gray!46}0.85                      & \cellcolor{gray!21}0.67                      & \cellcolor{gray!28}0.76                      & \cellcolor{gray!35}0.77                      & \cellcolor{gray!35}0.88                      & \cellcolor{gray!25}0.81                      & \cellcolor{gray!42}0.88                      & 6.11          \\
 & Electra                                      & \cellcolor{gray!42}0.84                      & \cellcolor{gray!21}0.64                      & \cellcolor{gray!60}\textbf{0.88}             & \cellcolor{gray!60}\textbf{0.74}             & \cellcolor{gray!42}0.77                      & \cellcolor{gray!35}0.77                      & \cellcolor{gray!56}0.94                      & \cellcolor{gray!15}0.78                      & \cellcolor{gray!15}0.86                      & 6.11          \\
 & DistilBERT                                   & \cellcolor{gray!15}0.82                      & \cellcolor{gray!35}0.67                      & \cellcolor{gray!49}0.86                      & \cellcolor{gray!39}0.70                      & \cellcolor{gray!28}0.76                      & \cellcolor{gray!60}\textbf{0.82}             & \cellcolor{gray!25}0.85                      & \cellcolor{gray!60}\textbf{0.83}             & \cellcolor{gray!15}0.86                      & 6.89          \\
 & HiCE                                         & \cellcolor{gray!28}0.83                      & \cellcolor{gray!35}0.67                      & \cellcolor{gray!28}0.80                      & \cellcolor{gray!53}0.72                      & \cellcolor{gray!15}0.75                      & \cellcolor{gray!49}0.78                      & \cellcolor{gray!18}0.81                      & \cellcolor{gray!35}0.82                      & \cellcolor{gray!42}0.88                      & 7.11          \\
 & RoBERTa                                      & \cellcolor{gray!15}0.82                      & \cellcolor{gray!35}0.67                      & \cellcolor{gray!21}0.78                      & \cellcolor{gray!39}0.70                      & \cellcolor{gray!28}0.76                      & \cellcolor{gray!25}0.76                      & \cellcolor{gray!46}0.89                      & \cellcolor{gray!35}0.82                      & \cellcolor{gray!42}0.88                      & 7.56          \\
 & HiCE (context)                               & \cellcolor{gray!28}0.83                      & \cellcolor{gray!60}\textbf{0.71}             & \cellcolor{gray!56}0.87                      & \cellcolor{gray!15}0.64                      & \cellcolor{gray!42}0.77                      & \cellcolor{gray!25}0.76                      & \cellcolor{gray!15}0.68                      & \cellcolor{gray!21}0.80                      & \cellcolor{gray!15}0.86                      & 8.44          \\
 \midrule

	\end{tabular}
}
\subtable[][\label{tab:results_twitter_1ArtificialOOV_stat}Statistics of the OOVs.]{
\begin{tabular}{c|m{1.9cm}ccccccccc} 

\toprule

\multicolumn{1}{c}{} &  & Archeage & HCR & Hobbit & IPhone6 & OMD & SS-Tweet & STS-Test & Sanders & UMICH \\ \midrule 

\multicolumn{1}{c}{} & \% of docs with OOV & 96.0 & 92.0 & 93.0 & 94.0 & 95.0 & 92.0 & 94.0 & 93.0 & 94.0 \\ \midrule 

\multirow{9}{*}{\rotatebox{90}{\makecell{\% of docs \\treated by \\the models}}} 

& Comick                                 & 100.0          & 100.0          & 100.0          & 100.0          & 100.0          & 100.0          & 100.0          & 100.0          & 100.0          \\ 
& Electra                                & 100.0          & 100.0          & 100.0          & 100.0          & 97.0          & 99.0          & 99.0          & 100.0          & 100.0          \\
& HiCE                                   & 100.0          & 100.0          & 100.0          & 100.0          & 100.0          & 100.0          & 100.0          & 100.0          & 100.0          \\
& HiCE (context)                         & 100.0          & 100.0          & 100.0          & 100.0          & 100.0          & 100.0          & 100.0          & 100.0          & 100.0          \\

& LSTM                                   & 95.0          & 96.0          & 93.0          & 93.0          & 95.0          & 95.0          & 92.0          & 95.0          & 93.0          \\

& Transformer                            & 95.0          & 96.0          & 93.0          & 93.0          & 95.0          & 95.0          & 92.0          & 95.0          & 93.0          \\
& GPT2                                   & 95.0          & 96.0          & 93.0          & 93.0          & 94.0          & 95.0          & 92.0          & 95.0          & 93.0          \\
& DistilBERT                             & 83.0          & 72.0          & 92.0          & 85.0          & 80.0          & 89.0          & 87.0          & 77.0          & 84.0          \\
& RoBERTa                                & 19.0          & 18.0          & 13.0          & 18.0          & 25.0          & 16.0          & 19.0          & 20.0          & 24.0          \\

 \midrule

	\end{tabular}}
\end{table}

\begin{table}[!htb]
\centering
\scriptsize
\setlength{\tabcolsep}{6pt} 
\renewcommand*{\arraystretch}{0.92}
\caption{Experiments with two artificial OOVs in each message.}%
    \label{tab:results_stat_twitter_2artificialOOVs}

\subtable[][\label{tab:results_twitter_2ArtificialOOV}Macro F-measure  obtained  on text categorization.]{
	\begin{tabular}{llcccccccccc}
		\toprule

& & \multicolumn{9}{c}{Macro F-measure} & \multirow{2}{*}{\makecell{Avg. \\ranking}} \\ \cmidrule(lr){3-11} 
& & Archeage & HCR & Hobbit & IPhone6 & OMD & SS-Tweet & STS-Test & Sanders & UMICH & \\  \cmidrule(lr){3-11}  \cmidrule(lr){12-12}

\multirow{5}{*}{\rotatebox{90}{Baselines}}   

 & Sum                                          & \cellcolor{gray!60}\textbf{0.84}             & \cellcolor{gray!60}\textbf{0.71}             & \cellcolor{gray!28}0.77                      & \cellcolor{gray!56}0.71                      & \cellcolor{gray!35}0.75                      & \cellcolor{gray!49}0.77                      & \cellcolor{gray!56}0.88                      & \cellcolor{gray!42}0.79                      & \cellcolor{gray!35}0.81                      & 4.00          \\
 & Average                                      & \cellcolor{gray!35}0.82                      & \cellcolor{gray!56}0.70                      & \cellcolor{gray!49}0.81                      & \cellcolor{gray!28}0.67                      & \cellcolor{gray!15}0.72                      & \cellcolor{gray!42}0.76                      & \cellcolor{gray!35}0.86                      & \cellcolor{gray!49}0.82                      & \cellcolor{gray!15}0.73                      & 6.78          \\
 & Random                                       & \cellcolor{gray!35}0.82                      & \cellcolor{gray!28}0.66                      & \cellcolor{gray!42}0.79                      & \cellcolor{gray!21}0.66                      & \cellcolor{gray!35}0.75                      & \cellcolor{gray!28}0.75                      & \cellcolor{gray!28}0.83                      & \cellcolor{gray!42}0.79                      & \cellcolor{gray!42}0.82                      & 7.44          \\
 & FastText                                     & \cellcolor{gray!35}0.82                      & \cellcolor{gray!35}0.67                      & \cellcolor{gray!39}0.78                      & \cellcolor{gray!15}0.64                      & \cellcolor{gray!18}0.73                      & \cellcolor{gray!28}0.75                      & \cellcolor{gray!46}0.87                      & \cellcolor{gray!28}0.78                      & \cellcolor{gray!60}\textbf{0.85}             & 7.56          \\
 & Zero                                         & \cellcolor{gray!35}0.82                      & \cellcolor{gray!49}0.69                      & \cellcolor{gray!28}0.77                      & \cellcolor{gray!18}0.65                      & \cellcolor{gray!60}\textbf{0.76}             & \cellcolor{gray!60}\textbf{0.78}             & \cellcolor{gray!15}0.76                      & \cellcolor{gray!18}0.76                      & \cellcolor{gray!28}0.78                      & 7.56                  \\
 
   \cmidrule(lr){3-11}  \cmidrule(lr){12-12}

\multirow{9}{*}{\rotatebox{90}{Deep learning}}    

 & LSTM                                         & \cellcolor{gray!60}\textbf{0.84}             & \cellcolor{gray!21}0.65                      & \cellcolor{gray!49}0.81                      & \cellcolor{gray!56}0.71                      & \cellcolor{gray!35}0.75                      & \cellcolor{gray!49}0.77                      & \cellcolor{gray!28}0.83                      & \cellcolor{gray!49}0.82                      & \cellcolor{gray!49}0.83                      & 4.56          \\
 & Electra                                      & \cellcolor{gray!35}0.82                      & \cellcolor{gray!15}0.64                      & \cellcolor{gray!60}\textbf{0.85}             & \cellcolor{gray!28}0.67                      & \cellcolor{gray!60}\textbf{0.76}             & \cellcolor{gray!60}\textbf{0.78}             & \cellcolor{gray!35}0.86                      & \cellcolor{gray!49}0.82                      & \cellcolor{gray!21}0.76                      & 5.44          \\
 & Comick                                       & \cellcolor{gray!35}0.82                      & \cellcolor{gray!21}0.65                      & \cellcolor{gray!53}0.82                      & \cellcolor{gray!28}0.67                      & \cellcolor{gray!60}\textbf{0.76}             & \cellcolor{gray!18}0.74                      & \cellcolor{gray!56}0.88                      & \cellcolor{gray!28}0.78                      & \cellcolor{gray!49}0.83                      & 6.00          \\
 & DistilBERT                                   & \cellcolor{gray!35}0.82                      & \cellcolor{gray!35}0.67                      & \cellcolor{gray!21}0.76                      & \cellcolor{gray!60}\textbf{0.72}             & \cellcolor{gray!21}0.74                      & \cellcolor{gray!42}0.76                      & \cellcolor{gray!46}0.87                      & \cellcolor{gray!49}0.82                      & \cellcolor{gray!18}0.75                      & 6.67          \\
 & GPT2                                         & \cellcolor{gray!49}0.83                      & \cellcolor{gray!28}0.66                      & \cellcolor{gray!56}0.84                      & \cellcolor{gray!39}0.68                      & \cellcolor{gray!35}0.75                      & \cellcolor{gray!28}0.75                      & \cellcolor{gray!21}0.81                      & \cellcolor{gray!28}0.78                      & \cellcolor{gray!42}0.82                      & 6.78          \\
 & Transformer                                  & \cellcolor{gray!18}0.79                      & \cellcolor{gray!49}0.69                      & \cellcolor{gray!28}0.77                      & \cellcolor{gray!28}0.67                      & \cellcolor{gray!60}\textbf{0.76}             & \cellcolor{gray!42}0.76                      & \cellcolor{gray!46}0.87                      & \cellcolor{gray!28}0.78                      & \cellcolor{gray!21}0.76                      & 6.78          \\
 & HiCE (context)                               & \cellcolor{gray!35}0.82                      & \cellcolor{gray!49}0.69                      & \cellcolor{gray!15}0.73                      & \cellcolor{gray!49}0.70                      & \cellcolor{gray!21}0.74                      & \cellcolor{gray!28}0.75                      & \cellcolor{gray!60}\textbf{0.90}             & \cellcolor{gray!15}0.72                      & \cellcolor{gray!42}0.82                      & 7.22          \\
 & HiCE                                         & \cellcolor{gray!15}0.78                      & \cellcolor{gray!15}0.64                      & \cellcolor{gray!28}0.77                      & \cellcolor{gray!42}0.69                      & \cellcolor{gray!35}0.75                      & \cellcolor{gray!42}0.76                      & \cellcolor{gray!18}0.80                      & \cellcolor{gray!60}\textbf{0.83}             & \cellcolor{gray!56}0.84                      & 7.33          \\
 & RoBERTa                                      & \cellcolor{gray!60}\textbf{0.84}             & \cellcolor{gray!49}0.69                      & \cellcolor{gray!15}0.73                      & \cellcolor{gray!42}0.69                      & \cellcolor{gray!35}0.75                      & \cellcolor{gray!15}0.73                      & \cellcolor{gray!32}0.84                      & \cellcolor{gray!28}0.78                      & \cellcolor{gray!32}0.80                      & 7.44          \\
 
 \midrule

	\end{tabular}
}
\subtable[][\label{tab:results_twitter_2ArtificialOOV_stat}Statistics of the OOVs.]{
\begin{tabular}{c|m{1.9cm}ccccccccc} 

\toprule

\multicolumn{1}{c}{} &  & Archeage & HCR & Hobbit & IPhone6 & OMD & SS-Tweet & STS-Test & Sanders & UMICH \\ \midrule 

\multicolumn{1}{c}{} & \% of docs with OOV & 99.0 & 100.0 & 100.0 & 100.0 & 100.0 & 99.0 & 100.0 & 99.0 & 100.0 \\ \midrule 

\multirow{9}{*}{\rotatebox{90}{\makecell{\% of docs \\treated by \\the models}}}

& Comick                                 & 100.0          & 100.0          & 100.0          & 100.0          & 100.0          & 100.0          & 100.0          & 100.0          & 100.0          \\ 

& HiCE                                   & 100.0          & 100.0          & 100.0          & 100.0          & 100.0          & 100.0          & 100.0          & 100.0          & 100.0          \\
& HiCE (context)                         & 100.0          & 100.0          & 100.0          & 100.0          & 100.0          & 100.0          & 100.0          & 100.0          & 100.0          \\

& Electra                                & 100.0          & 100.0          & 100.0          & 100.0          & 100.0          & 100.0          & 99.0          & 100.0          & 100.0          \\

& GPT2                                   & 99.0          & 99.0          & 99.0          & 98.0          & 100.0          & 99.0          & 99.0          & 100.0          & 99.0          \\

& LSTM                                   & 98.0          & 98.0          & 98.0          & 97.0          & 99.0          & 99.0          & 98.0          & 99.0          & 96.0          \\
& Transformer                            & 98.0          & 98.0          & 98.0          & 97.0          & 99.0          & 99.0          & 98.0          & 99.0          & 96.0          \\
& DistilBERT                             & 95.0          & 88.0          & 98.0          & 90.0          & 92.0          & 95.0          & 94.0          & 89.0          & 92.0          \\
& RoBERTa                                & 30.0          & 33.0          & 26.0          & 31.0          & 44.0          & 29.0          & 35.0          & 36.0          & 46.0          \\

\midrule

	\end{tabular}}
\end{table}

\begin{table}[!htb]
\centering
\scriptsize
\setlength{\tabcolsep}{6pt} 
\renewcommand*{\arraystretch}{0.92}
\caption{Experiments with three artificial OOV in each message.}%
    \label{tab:results_stat_twitter_3artificialOOVs}

\subtable[][\label{tab:results_twitter_3ArtificialOOV}Macro F-measure  obtained  on text categorization.]{
	\begin{tabular}{llcccccccccc}
		\toprule

& & \multicolumn{9}{c}{Macro F-measure} & \multirow{2}{*}{\makecell{Avg. \\ranking}} \\ \cmidrule(lr){3-11} 
& & Archeage & HCR & Hobbit & IPhone6 & OMD & SS-Tweet & STS-Test & Sanders & UMICH & \\  \cmidrule(lr){3-11}  \cmidrule(lr){12-12}

\multirow{5}{*}{\rotatebox{90}{Baselines}}   

 & Average                                      & \cellcolor{gray!35}0.78                      & \cellcolor{gray!42}0.66                      & \cellcolor{gray!56}0.80                      & \cellcolor{gray!60}\textbf{0.74}             & \cellcolor{gray!60}\textbf{0.73}             & \cellcolor{gray!60}\textbf{0.76}             & \cellcolor{gray!60}\textbf{0.89}             & \cellcolor{gray!35}0.78                      & \cellcolor{gray!39}0.77                      & 3.33          \\
 & Random                                       & \cellcolor{gray!56}0.81                      & \cellcolor{gray!35}0.65                      & \cellcolor{gray!21}0.72                      & \cellcolor{gray!35}0.70                      & \cellcolor{gray!49}0.72                      & \cellcolor{gray!39}0.75                      & \cellcolor{gray!60}\textbf{0.89}             & \cellcolor{gray!49}0.79                      & \cellcolor{gray!49}0.79                      & 5.00          \\
 & Sum                                          & \cellcolor{gray!56}0.81                      & \cellcolor{gray!42}0.66                      & \cellcolor{gray!42}0.77                      & \cellcolor{gray!35}0.70                      & \cellcolor{gray!28}0.70                      & \cellcolor{gray!39}0.75                      & \cellcolor{gray!35}0.85                      & \cellcolor{gray!60}\textbf{0.81}             & \cellcolor{gray!39}0.77                      & 5.22          \\
 & Zero                                         & \cellcolor{gray!49}0.80                      & \cellcolor{gray!28}0.64                      & \cellcolor{gray!42}0.77                      & \cellcolor{gray!42}0.71                      & \cellcolor{gray!39}0.71                      & \cellcolor{gray!60}\textbf{0.76}             & \cellcolor{gray!28}0.84                      & \cellcolor{gray!15}0.42                      & \cellcolor{gray!46}0.78                      & 6.56          \\
 & FastText                                     & \cellcolor{gray!28}0.77                      & \cellcolor{gray!53}0.68                      & \cellcolor{gray!15}0.61                      & \cellcolor{gray!28}0.69                      & \cellcolor{gray!49}0.72                      & \cellcolor{gray!28}0.73                      & \cellcolor{gray!18}0.81                      & \cellcolor{gray!18}0.73                      & \cellcolor{gray!60}\textbf{0.81}             & 8.33                    \\

  \cmidrule(lr){3-11}  \cmidrule(lr){12-12}

\multirow{9}{*}{\rotatebox{90}{Deep learning}} 

& Comick                                       & \cellcolor{gray!60}\textbf{0.82}             & \cellcolor{gray!60}\textbf{0.69}             & \cellcolor{gray!53}0.78                      & \cellcolor{gray!21}0.68                      & \cellcolor{gray!49}0.72                      & \cellcolor{gray!39}0.75                      & \cellcolor{gray!49}0.88                      & \cellcolor{gray!35}0.78                      & \cellcolor{gray!21}0.74                      & 5.00          \\
 & Electra                                      & \cellcolor{gray!49}0.80                      & \cellcolor{gray!35}0.65                      & \cellcolor{gray!42}0.77                      & \cellcolor{gray!60}\textbf{0.74}             & \cellcolor{gray!39}0.71                      & \cellcolor{gray!39}0.75                      & \cellcolor{gray!60}\textbf{0.89}             & \cellcolor{gray!35}0.78                      & \cellcolor{gray!15}0.71                      & 5.44          \\
 & Transformer                                  & \cellcolor{gray!15}0.76                      & \cellcolor{gray!49}0.67                      & \cellcolor{gray!32}0.76                      & \cellcolor{gray!42}0.71                      & \cellcolor{gray!49}0.72                      & \cellcolor{gray!25}0.72                      & \cellcolor{gray!49}0.88                      & \cellcolor{gray!49}0.79                      & \cellcolor{gray!56}0.80                      & 5.78          \\
 & GPT2                                         & \cellcolor{gray!42}0.79                      & \cellcolor{gray!42}0.66                      & \cellcolor{gray!60}\textbf{0.83}             & \cellcolor{gray!18}0.67                      & \cellcolor{gray!28}0.70                      & \cellcolor{gray!21}0.71                      & \cellcolor{gray!42}0.87                      & \cellcolor{gray!49}0.79                      & \cellcolor{gray!32}0.76                      & 7.11          \\
 & DistilBERT                                   & \cellcolor{gray!28}0.77                      & \cellcolor{gray!15}0.61                      & \cellcolor{gray!28}0.75                      & \cellcolor{gray!53}0.73                      & \cellcolor{gray!28}0.70                      & \cellcolor{gray!60}\textbf{0.76}             & \cellcolor{gray!21}0.82                      & \cellcolor{gray!56}0.80                      & \cellcolor{gray!28}0.75                      & 7.67          \\
 & LSTM                                         & \cellcolor{gray!15}0.76                      & \cellcolor{gray!28}0.64                      & \cellcolor{gray!21}0.72                      & \cellcolor{gray!42}0.71                      & \cellcolor{gray!15}0.63                      & \cellcolor{gray!39}0.75                      & \cellcolor{gray!28}0.84                      & \cellcolor{gray!35}0.78                      & \cellcolor{gray!56}0.80                      & 8.44          \\
 & RoBERTa                                      & \cellcolor{gray!15}0.76                      & \cellcolor{gray!21}0.63                      & \cellcolor{gray!42}0.77                      & \cellcolor{gray!28}0.69                      & \cellcolor{gray!39}0.71                      & \cellcolor{gray!60}\textbf{0.76}             & \cellcolor{gray!15}0.78                      & \cellcolor{gray!35}0.78                      & \cellcolor{gray!21}0.74                      & 8.44          \\
 & HiCE                                         & \cellcolor{gray!35}0.78                      & \cellcolor{gray!60}\textbf{0.69}             & \cellcolor{gray!25}0.74                      & \cellcolor{gray!15}0.65                      & \cellcolor{gray!18}0.69                      & \cellcolor{gray!15}0.67                      & \cellcolor{gray!35}0.85                      & \cellcolor{gray!35}0.78                      & \cellcolor{gray!39}0.77                      & 8.78          \\
 & HiCE (context)                               & \cellcolor{gray!28}0.77                      & \cellcolor{gray!15}0.61                      & \cellcolor{gray!42}0.77                      & \cellcolor{gray!49}0.72                      & \cellcolor{gray!28}0.70                      & \cellcolor{gray!18}0.70                      & \cellcolor{gray!35}0.85                      & \cellcolor{gray!21}0.76                      & \cellcolor{gray!18}0.72                      & 9.33          \\
 \midrule

	\end{tabular}
}
\subtable[][\label{tab:results_twitter_3ArtificialOOV_stat}Statistics of the OOVs.]{
\begin{tabular}{c|m{1.9cm}ccccccccc} 

\toprule

\multicolumn{1}{c}{} &  & Archeage & HCR & Hobbit & IPhone6 & OMD & SS-Tweet & STS-Test & Sanders & UMICH \\ \midrule 

\multicolumn{1}{c}{} & \% of docs with OOV & 99.0 & 100.0 & 100.0 & 100.0 & 100.0 & 99.0 & 100.0 & 99.0 & 100.0 \\ \midrule

\multirow{9}{*}{\rotatebox{90}{\makecell{\% of docs \\treated by \\the models}}} 

& Comick                                 & 100.0          & 100.0          & 100.0          & 100.0          & 100.0          & 100.0          & 100.0          & 100.0          & 100.0          \\ 

& Electra                                & 100.0          & 100.0          & 100.0          & 100.0          & 100.0          & 100.0          & 100.0          & 100.0          & 100.0          \\
& HiCE                                   & 100.0          & 100.0          & 100.0          & 100.0          & 100.0          & 100.0          & 100.0          & 100.0          & 100.0          \\
& HiCE (context)                         & 100.0          & 100.0          & 100.0          & 100.0          & 100.0          & 100.0          & 100.0          & 100.0          & 100.0          \\

& LSTM                                   & 99.0          & 99.0          & 99.0          & 98.0          & 100.0          & 99.0          & 98.0          & 100.0          & 97.0          \\
& Transformer                            & 99.0          & 99.0          & 99.0          & 98.0          & 100.0          & 99.0          & 98.0          & 100.0          & 97.0          \\
& GPT2                                   & 99.0          & 100.0          & 99.0          & 99.0          & 100.0          & 100.0          & 100.0          & 100.0          & 100.0          \\
& DistilBERT                             & 97.0          & 94.0          & 99.0          & 94.0          & 96.0          & 97.0          & 96.0          & 93.0          & 94.0          \\
& RoBERTa                                & 42.0          & 44.0          & 42.0          & 43.0          & 60.0          & 42.0          & 49.0          & 50.0          & 61.0          \\

 \midrule

	\end{tabular}}
\end{table}

In the experiments where one artificial OOV was inserted in each message, Average obtained the best macro F-measure in two datasets (STS-test and Sanders) and the best average ranking. Transformer was the best DL model and obtained the third best average ranking. 

In the experiments with two artificial OOVs, Average and Transformer did not achieve the same success as in experiments with one artificial OOV. In this case, Average did not obtain the best result in any dataset, despite having obtained the second best average ranking among the baseline methods. Moreover, the average ranking obtained by Transformer was only the sixth best among the DL methods. In this experiment, the best average ranking was obtained by LSTM. 

In the experiment with three online OOVs, Average obtained the best average ranking, and Random and Comick tied with the second best average ranking. Although the best DL method was different in each experiment with artificial OOVs, Comick and Electra obtained consistent results, being among the four best DL models in terms of average ranking in all of them. Another method that performed similarly in all three experiments were HiCE and HiCE (context), but in a negative way, since they obtained one of the three worst average rankings in all of them.

If we analyze the statistics of OOVs treated in the experiments with artificial OOVs (Tables \ref{tab:results_twitter_1ArtificialOOV_stat}, \ref{tab:results_twitter_2ArtificialOOV_stat}, and \ref{tab:results_twitter_3ArtificialOOV_stat}), we can see that, as in the experiments with the original messages, Roberta was the DL model that handled the lowest percentage of documents with OOVs. This was reflected in the results obtained by Roberta, as it achieved one of the four worst average rankings among the DL models in all experiments with artificial OOVs.

\subsection{NER and POS tagging}

We performed experiments with three datasets commonly used in studies that address OOVs \cite{hu:2019_hice}: 
\begin{itemize}[topsep=2.5pt]
	\item \textbf{Rare-NER} \cite{derczynski:2017-RareNER}: a NER dataset with unusual and unseen entities in the context of emerging discussions.
	
	\item \textbf{Bio-NER} \cite{kim:2004-BioNER}: a NER dataset with sentences that have technical terms in the biology domain. 
	
	\item \textbf{Twitter-POS} \cite{ritter:2011-nerTwitter}: a POS tagging dataset of Twitter messages. 
\end{itemize}

All sentences were converted to lowercase. All documents from all datasets are already provided separated by terms and each term has an associated label (entity or POS). Therefore, it was not necessary to perform tokenization.

Table \ref{tab:datasets_NER_PosTagging} presents the main statistics about the datasets, where $|D|$ is the number of documents, and $|V|$ is the number of unique terms (vocabulary). Moreover,  $\mathcal{M}^{t}$ and $\mathcal{I}^{t}$ are the median and the interquartile range of the number of terms per document.

 \begin{table}[!htb]
 	\centering
 	\footnotesize
 	\caption{Basic statistics about the datasets for NER and POS tagging}\label{tab:datasets_NER_PosTagging}
 	\setlength{\tabcolsep}{6.5pt} 
 	\begin{tabular}{lllll}
 		\toprule
 		Dataset  & $D$   & $|V|$ & $\mathcal{M}^{t}$ & $\mathcal{I}^{t}$ \\ \midrule
 		Rare-NER       & 5,690  & 20,773 & 17 & 13 \\
 		Bio-NER        & 22,402 & 25,103 & 24 & 15 \\
 		Twitter-POS  & 787    & 4,766 & 20 & 12 \\ \midrule
 	\end{tabular}
 \end{table}

\subsubsection{Evaluation}

The experiments with NER and POS tagging were carried out using a Bidirectional LSTM with a time distributed dense layer built using Keras  on top of TensorFlow. All the documents were padded or truncated to 200 words. The OOVs that were not handled and paddings words were represented by a vector of zeros. 

The experiments were performed using a holdout validation with 70\% of the documents in the training set and 30\% in the test set. To compare the results in the NER task, we employed the entity level F-measure. In the experiments with POS-tagging, we evaluated the prediction performance using the token accuracy.

\subsubsection{Results}
\label{sec:results_NER_POSTagging}

Table \ref{tab:results_ner_POSTagging} presents the results obtained. For each dataset, the method that obtained the best F-measure (NER) or the best accuracy (POS-tagging), obtained rank 1, while the worst method obtained rank 14. The methods are
sorted by the average ranking. In addition, Table \ref{tab:estatisticas_OOV_NER_PosTagging} presents some statistics about the OOVs of the evaluated datasets.

\begin{table}[!htb]
	\centering
	\scriptsize
	\setlength{\tabcolsep}{4.5pt} 
	\caption{Experiments on NER and POS tagging.}
	\label{tab:experiments_ner_POSTagging}
\subtable[][\label{tab:results_ner_POSTagging}Performance on NER and POS tagging.]{
	\begin{tabular}{llcccc}
		\toprule
		
& & \multicolumn{2}{c}{\makecell{NER \\(F-measure)}} & \makecell{POS tagging\\ (accuracy)} & \multirow{2}{*}{\makecell{Avg. \\ranking}} \\ \cmidrule(lr){3-4} \cmidrule(lr){5-5} 
& & \makecell{Bio-\\NER} & \makecell{Rare-\\NER} & \makecell{Twitter-\\POS} \\ \cmidrule(lr){3-4} \cmidrule(lr){5-5} \cmidrule(lr){6-6} 

\multirow{5}{*}{\rotatebox{90}{Baselines}} & FastText                                     & \cellcolor{gray!60}\textbf{0.67}             & \cellcolor{gray!60}\textbf{0.46}             & \cellcolor{gray!60}\textbf{0.78}             & 1.00          \\
 & Average                                      & \cellcolor{gray!56}0.63                      & \cellcolor{gray!60}\textbf{0.46}             & \cellcolor{gray!56}0.76                      & 1.67          \\
 & Random                                       & \cellcolor{gray!53}0.62                      & \cellcolor{gray!49}0.45                      & \cellcolor{gray!35}0.74                      & 5.00          \\
 & Zero                                         & \cellcolor{gray!53}0.62                      & \cellcolor{gray!15}0.37                      & \cellcolor{gray!56}0.76                      & 6.33          \\
 & Sum                                          & \cellcolor{gray!35}0.60                      & \cellcolor{gray!42}0.44                      & \cellcolor{gray!35}0.74                      & 7.33          \\
 \cmidrule(lr){3-4} \cmidrule(lr){5-5} \cmidrule(lr){6-6}

\multirow{9}{*}{\rotatebox{90}{Deep learning}} & LSTM                                         & \cellcolor{gray!32}0.59                      & \cellcolor{gray!60}\textbf{0.46}             & \cellcolor{gray!46}0.75                      & 5.00          \\
 & Comick                                       & \cellcolor{gray!53}0.62                      & \cellcolor{gray!42}0.44                      & \cellcolor{gray!25}0.73                      & 6.67          \\
 & RoBERTa                                      & \cellcolor{gray!53}0.62                      & \cellcolor{gray!35}0.42                      & \cellcolor{gray!25}0.73                      & 7.33          \\
 & Transformer                                  & \cellcolor{gray!15}0.56                      & \cellcolor{gray!49}0.45                      & \cellcolor{gray!46}0.75                      & 7.67          \\
 & DistilBERT                                   & \cellcolor{gray!28}0.58                      & \cellcolor{gray!21}0.39                      & \cellcolor{gray!56}0.76                      & 8.00          \\
 & Electra                                      & \cellcolor{gray!21}0.57                      & \cellcolor{gray!25}0.41                      & \cellcolor{gray!46}0.75                      & 9.33          \\
 & GPT2                                         & \cellcolor{gray!21}0.57                      & \cellcolor{gray!35}0.42                      & \cellcolor{gray!35}0.74                      & 9.33          \\
 & HiCE (context)                               & \cellcolor{gray!39}0.61                      & \cellcolor{gray!21}0.39                      & \cellcolor{gray!25}0.73                      & 10.00          \\
 & HiCE                                         & \cellcolor{gray!28}0.58                      & \cellcolor{gray!35}0.42                      & \cellcolor{gray!15}0.71                      & 10.67                   \\ \midrule

	\end{tabular}}
	\qquad
\subtable[][\label{tab:estatisticas_OOV_NER_PosTagging}Statistics of the OOVs.]{	
\begin{tabular}{c|m{1.9cm}ccc} 

\toprule

\multicolumn{1}{c}{} &  & \makecell{Bio-\\NER} & \makecell{Rare\\-NER} & \makecell{Twitter\\-POS} \\ \midrule 
\multicolumn{1}{c}{} & \% of docs with OOV & 85.0 & 69.0 & 85.0 \\ \midrule 

\multirow{9}{*}{\rotatebox{90}{\makecell{\% of docs treated \\by the models}}} 
& Comick                                 & 100.0          & 100.0          & 100.0          \\
& HiCE                                   & 100.0          & 100.0          & 100.0          \\
& HiCE (context)                         & 100.0          & 100.0          & 100.0          \\

& GPT2                                   & 98.0          & 92.0          & 90.0          \\

& LSTM                                   & 98.0          & 91.0          & 89.0          \\
& Transformer                            & 98.0          & 91.0          & 89.0          \\

& DistilBERT                             & 99.0          & 84.0          & 85.0          \\

& Electra                                & 93.0          & 82.0          & 85.0          \\

& RoBERTa                                & 23.0          & 67.0          & 74.0          \\
 \midrule
	\end{tabular}
	}
\end{table}

FastText was the best technique in the experiments with NER and POS tagging. 
In the experiment with Bio-NER, the F-measure obtained by FastText was 5\% higher than the second best score (Average) and 20\% higher than the lowest score (Transformer). Furthermore, the average ranking of simple baselines techniques (Sum, Zero, Average, and Random) was higher than most DL methods. The reason may be that OOV words have little impact on the text of the evaluated datasets, for being rare words or have little semantic importance. We also believe that the noise from the documents in T7 (Twitter messages) may have affected the quality of the embeddings.

Among the DL models, LSTM obtained the best average ranking. Moreover, Comick repeated the good performance shown in the text categorization task, obtaining the second best average ranking in NER and POS tasks. This model, unlike most other DL models (except Hice) that we have evaluated, in addition to the context, analyze the morphological structure of the OOVs. The results indicate that this characteristic may have benefited Comick in relation to the other DL models.

By the statistics shown in Table \ref{tab:estatisticas_OOV_NER_PosTagging}, we can note that the NER and POS tagging datasets have more OOVs than the text categorization datasets (Table \ref{tab:results_twitter_stat}). In addition, NER and POS tagging can be more impacted by OOVs than the text categorization task, as almost all words are associated with a label (entity or POS). As in experiments with the text classification task, in the experiments with NER and POS tagging, RoBERTa  was the DL model that treated the lowest percentage of documents with OOVS. 

As in experiments with the text classification task, in the experiments with NER and POS tagging, RoBERTa  was the DL model that treated the lowest percentage of documents with OOVS. This probably influenced the low performance it obtained on the Rare-NER and Twitter-POS datasets.

DistilBERT, the best method in the intrinsic evaluation (Section \ref{sec:results_chimera}) and the second worst method in the text categorization  (Section \ref{sec:results_twitter}), obtained the second best performance in the POS tagging task, but was one of the worst methods in the experiments with NER.

\section{Conclusions}
\label{sec:conclusao}

The phenomenon of OOVs is a major problem in natural language processing tasks. Documents that have OOVs are usually incompletely represented by distributed text representation models. The lack of one or more words can significantly change the semantics of a sentence.

Distributed text representation models are not incremental and, therefore, the training process is performed only once due to the high computational cost demanded. As the model generated in the training is not updated over time, it is unable to deal with new words that were not seen during its training. Therefore, the more dynamic the communication becomes, the more OOVs appear, and the faster the model becomes obsolete.

In this paper, we presented a comprehensive performance evaluation of different DL models applied to handle OOVs. Among the evaluated models, DistilBERT, GPT2, Electra, LSTM, Transformer, and Roberta infer the embedding for a given OOV using approximation by the terms that appear next to the OOV in the sentence. Comick and HiCE, in addition to the context, use the morphological structure of the OOV for the inference. 

To analyze these models, we performed an intrinsic evaluation using the benchmark Chimera dataset. We also performed an extrinsic evaluation with the text categorization task using nine public and well-known datasets for opinion polarity detection on Twitter messages, and with the tasks of NER and POS tagging, using three datasets frequently used in related studies.

There was no model that obtained the best performance in all evaluations. However, in general, Comick obtained a good performance in all extrinsic evaluation tasks, which resulted in higher average ranking than most other evaluated DL models. The ability of this model to analyze the morphological structure of OOVs, in addition to their context, may have contributed to achieve a superior performance, although Hice did not achieve the same success even having the same characteristics.

Considering each experiment more specifically, in the intrinsic evaluation, DistilBERT obtained the best performance, with a significant difference to other methods. In the text categorization task, in general, Comick was the best method to infer embeddings for OOVs. Finally, in NER and POS tagging, the best performance was obtained by one of the baseline techniques: FastText. 

Based on the results, we can conclude that the task of inferring embeddings for OOVs generates different challenges for different evaluated scenarios. Therefore, we recommend that research on OOV handing techniques be addressed to specific tasks, increasing the chance of success. 

We also noticed that for some scenarios with noisy texts (as in the datasets with Twitter messages) or sentences full of technical terms (as in the Bio-NER dataset), the context of OOVs and their morphological structure may not be enough to infer a good embedding. Therefore, we recommend using an architecture for OOV handling that combines the techniques analyzed in this study with simpler techniques based on spell checker and semantic dictionaries.

%
%
%

\bibliographystyle{splncs04}

{%
\linespread{0.86}

}%

\end{document}